\documentclass[letterpaper]{article} 
\usepackage{aaai2026}  
\usepackage{times}  
\usepackage{helvet}  
\usepackage{courier}  
\usepackage[hyphens]{url}  
\usepackage{graphicx} 
\usepackage{natbib}  
\usepackage{caption} 
\usepackage{algorithm}
\usepackage{algpseudocode}
\usepackage{booktabs}       
\usepackage{amsfonts}       
\usepackage{nicefrac}       
\usepackage{microtype}      
\usepackage{xcolor}         
\usepackage{todonotes}

\usepackage{makecell}
\usepackage{subcaption}
\usepackage{float}
\usepackage{siunitx}
\usepackage[switch]{lineno}

\usepackage{amsmath}
\usepackage{mathrsfs}
\usepackage{tabularx}

\usepackage{tikz}
\usepackage{multirow}
\usepackage{adjustbox} 

\usepackage{newfloat}
\usepackage{listings}
\DeclareCaptionStyle{ruled}{labelfont=normalfont,labelsep=colon,strut=off} 
\floatstyle{ruled}
\newfloat{listing}{tb}{lst}{}
\floatname{listing}{Listing}
\title{HKT: A Biologically Inspired Framework for Modular Hereditary\\ Knowledge Transfer in Neural Networks}
\author{Yanick Chistian Tchenko\textsuperscript{\rm 1},
    Felix Mohr \textsuperscript{\rm 2},\\
    Hicham Hadj Abdelkader\textsuperscript{\rm 1},
    Hedi Tabia\textsuperscript{\rm 1}
}
\affiliations{
    \textsuperscript{\rm 1}University Paris Saclay, 9 Rue Joliot Curie, 91190 Gif-sur-Yvette, France\\

    \textsuperscript{\rm 2}Universidad de La Sabana, Km. 7, Autopista Norte de Bogotá. Chía, Cundinamarca, Colombia\\
    \{Yanickchristian.tchenko, Hicham.hadjabdelkader, Hedi.tabia\}@univ-evry.fr, Felix.mohr@unisabana.edu.co
}

\usepackage{xspace}
\newcommand{\mathsymbol}[2]{\newcommand{#1}{\ensuremath{\mathit{#2}}\xspace}}

\mathsymbol{\loss}{\mathscr{L}}

\mathsymbol{\parent}{\mathcal{P}}
\mathsymbol{\parentparams}{\theta_\parent}
\mathsymbol{\parentblock}{p}
\mathsymbol{\child}{\mathcal{C}}
\mathsymbol{\childparams}{\theta_\child}
\mathsymbol{\childblock}{c}

\mathsymbol{\cE}{\mathcal{E}}
\mathsymbol{\cT}{\mathcal{T}}
\mathsymbol{\cM}{\mathcal{M}}
\mathsymbol{\transcodedknowledge}{\tau}
\mathsymbol{\fusedknowledge}{\phi}

\newcommand{\circled}[1]{\tikz[baseline=(char.base)]{
            \node[shape=circle,draw,inner sep=1pt] (char) {#1};}}
            
\begin{document}

\maketitle

\begin{abstract}
A prevailing trend in neural network research suggests that model performance improves with increasing depth and capacity—often at the cost of integrability and efficiency. In this paper, we propose a strategy to optimize small, deployable models by enhancing their capabilities through structured knowledge inheritance. We introduce Hereditary Knowledge Transfer (HKT), a biologically inspired framework for modular and selective transfer of task-relevant features from a larger, pretrained parent network to a smaller child model. Unlike standard knowledge distillation, which enforces uniform imitation of teacher outputs, HKT draws inspiration from biological inheritance mechanisms—such as memory RNA transfer in planarians—to guide a multi-stage process of feature transfer. Neural network blocks are treated as functional carriers, and knowledge is transmitted through three biologically motivated components: Extraction, Transfer, and Mixture (ETM). A novel Genetic Attention (GA) mechanism governs the integration of inherited and native representations, ensuring both alignment and selectivity. We evaluate HKT across diverse vision tasks, including optical flow (Sintel, KITTI), image classification (CIFAR-10), and semantic segmentation (LiTS), demonstrating that it significantly improves child model performance while preserving its compactness. The results show that HKT consistently outperforms conventional distillation approaches, offering a general-purpose, interpretable, and scalable solution for deploying high-performance neural networks in resource-constrained environments. Our code is available under \url{https://github.com/christian-tchenko/HKT-ResNet.git}.
\end{abstract}

\section{Introduction}
\begin{figure}[ht!]
    \centering
    \includegraphics[width=\linewidth]{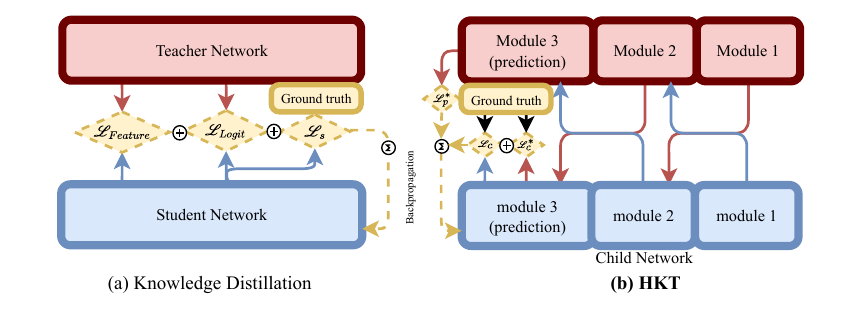}
    \vspace{-2\baselineskip} 
   \caption{\textbf{Comparing KD and HKT:} \textbf{(a) KD} – A teacher model guides a student model by sharing predictions and ground truth labels, optimizing the student through combined loss minimization. \textbf{(b) HKT} – A child model inherits selectively from both a small base model and a parent model, focusing on ground truth optimization while leveraging inherited and self-derived insights.}
    \label{fig:hkt_kd}
\end{figure}

Effective and accurate neural networks are critical for resource-constrained environments, such as edge deployment in natural language processing (NLP) and computer vision. These scenarios demand models that balance high performance with computational efficiency. A widely adopted approach to achieve this balance is \textit{Knowledge Distillation (KD)} \cite{hinton_distilling_2015}, where a smaller student model learns by mimicking the outputs of a larger teacher model. However, KD transfers knowledge indiscriminately, often propagating irrelevant or suboptimal teacher capabilities that can degrade student performance (see Figure \ref{fig:hkt_kd} (a)). This limitation is particularly evident in tasks that require fine-grained spatial understanding, such as optical flow estimation, where detailed motion analysis is essential. These challenges highlight the need for a targeted and task-specific knowledge transfer mechanism.
To address these limitations, this paper introduces \textit{Hereditary Knowledge Transfer (HKT)}, a biologically inspired framework designed for selective stage-wise inheritance of task-specific abilities (see Figure \ref{fig:hkt_kd} (b) ). Unlike KD, HKT employs a structured parent-child architecture that leverages the ETM triad (Extraction, Transformation, Mixture) to facilitate modular and targeted knowledge transfer. An enhanced Genetic Attention (GA) mechanism dynamically prioritizes task-relevant features, ensuring efficient merging between inherited and self-learned knowledge. 

Our contributions are as follows: (1) \textbf{A novel framework (HKT):} We propose HKT as a new paradigm for optimizing both the \emph{quantity} and \emph{quality} of transferred knowledge, inspired by hereditary inheritance. (2) \textbf{3-Stage HKT (3HKT):} We introduce a multi-stage design that enhances feature alignment and task-specific adaptability, particularly for high-dimensional, dynamic tasks such as optical flow estimation. (3) \textbf{Modularity and scalability:} HKT enables structured, functional, and granular transfer between aligned blocks in networks of arbitrary depth, making it flexible and generalizable. (4) \textbf{Extensive empirical validation:} Across diverse benchmarks (Sintel, KITTI, CIFAR-10, LiTS), HKT consistently outperforms state-of-the-art KD techniques in accuracy, efficiency, and generalization.

\section{Related Work}
\paragraph{Knowledge Transfer Frameworks ---}Knowledge transfer enables the efficient deployment of neural networks in resource-constrained environments. \textit{Knowledge Distillation (KD)} is a widely used method where smaller student models mimic teacher outputs. While effective for tasks like object detection \cite{wang2024crosskdcrossheadknowledgedistillation} and speech recognition, KD indiscriminately transfers both strengths and weaknesses, limiting its utility in tasks requiring fine-grained spatial understanding or adaptability. Recent approaches, such as DDFlow \cite{liu2019ddflowlearningopticalflow} and MDFlow \cite{9882137}, improve optical flow learning through data distillation and mutual distillation but remain constrained by KD’s reliance on indirect imitation. Hierarchical transfer models \cite{park2019relational} and layer-wise frameworks \cite{xu2021gmflow} enhance granularity but lack scalability for modular architectures. The proposed HKT framework introduces discriminative, modularized inheritance using a parent-child architecture.
\paragraph{Biological Inspiration in Neural Design ---}HKT is inspired by biological processes, particularly the mechanisms of genetic inheritance. In biology, information is transferred across generations through structured processes like DNA transcription and translation \cite{Alberts2002}. These processes involve messenger RNA (mRNA) conveying genetic instructions, transfer RNA (tRNA) delivering specific components, and ribosomal RNA (rRNA) synthesizing proteins to express traits. HKT mirrors these biological mechanisms using the ETM triad (Extraction, Transformation, Mixture), which facilitates selective, modular knowledge transfer. Similarly, GA mimics gene expression by dynamically prioritizing critical features during transfer, ensuring that inherited knowledge aligns with task-specific requirements. Prior work on bio-inspired neural architectures \cite{mcswiggin2024epigenetic,moelling2024epigenetics} highlights the potential of leveraging biological principles to enhance neural network efficiency and adaptability. The HKT framework extends these principles by integrating GA and ETM into a three-stage design, improving the granularity and modularity of neural inheritance.
\paragraph{Modular Design and Scalability ---}Scalability and modularity are essential for extending knowledge transfer frameworks beyond fixed architectures. Modular designs decompose networks into functional components, supporting multi-task and transfer learning. While pruning methods \cite{lu2023energy} and low-rank factorization \cite{DBLP:journals/corr/Yang15b} reduce model complexity effectively, they often lack the fine-grained alignment required for stage-wise transfer. HKT overcomes these limitations by introducing an additional inheritance stage, enabling granular feature alignment and dynamic balancing of inherited and self-learned features. This structured, multi-stage approach ensures consistent performance across components, establishing a scalable foundation for complex AI tasks.

\begin{figure*}[h!]
\centering
    \includegraphics[width= 1\linewidth]{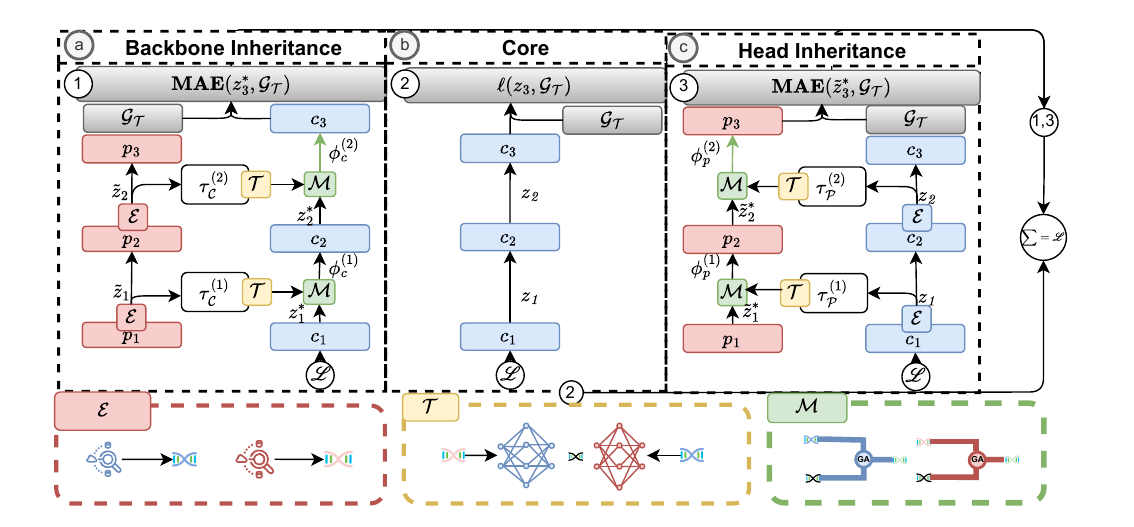}
    \vspace{-2\baselineskip} 
   \caption[3HKT]{\textbf{System Overview:} In HKT, knowledge is transferred from the parent network (red) to the child network (blue) through stages using the ETM triad. Modular and core losses are computed, aggregated, and backpropagated to optimize the child network.
   }
    \label{fig:hkt}
\end{figure*}

\section{Problem Statement}\label{sec:problem}

We address the standard task of finding the parameters $\theta_\child$ that minimize the expected loss for a deep network \child under a loss \loss in a given data domain, i.e.,
\begin{equation}
    \arg\min_{\theta_\child}~~\mathbb{E}_{(x, y)}~\loss(\child(x), y),
\end{equation}
where $\loss(\hat y, y)$ is the loss of predicting $\hat y$ when the true answer should be $y$ and both \child and \loss are differentiable.

The standard setting is only extended in that we provide a \emph{pre-trained} network \parent that was already trained on the same task and has a structural and semantic analogy to \child but is significantly more complex.
Formally, we assume that the layers of both \child and \parent are organized into $n$ functional blocks $\{\childblock_1,..,\childblock_n\}$ and $\{\parentblock_1,..,\parentblock_n\}$, each with a specific purpose (e.g., embedding, feature extraction) and that corresponding blocks $\childblock_i$ and $\parentblock_i$ are functionally aligned.
In other words, $c_i$ and $p_i$ are intended to assume the same semantic functionality in their context, but $c_i$ is significantly simpler than $p_i$.
\parent is referred to as the \emph{parent}, and \child as the \emph{child}.

The main difference to knowledge distillation (KD) is that the objective is still to learn the ground truth, so the child network does \emph{not} learn to mimic the parent network.
In other words, the parent network \parent is not part of the mathematical problem description but is simply an additional \emph{resource} that can be used to address that task.
Making use of this additional resource \parent might allow to find a better local optimum for $\theta_\child$ than would be found if \parent would not be available.

\section{Methodology}\label{sec:method}
Our approach to leverage the parent network \parent to train a competent child \child is to mix-in knowledge of the parent into the forward pass of the child using the triad we call the \emph{extract-transfer-mixture} mechanism.

\subsection{ETM Architecture \& Standard Pipeline}
The core of the HKT framework is a non-parametric structure, which we call the ETM architecture, which facilitates biologically inspired knowledge inheritance.
This architecture, which is depicted in Figure \ref{fig:hkt}, consists of three functional units:
(1) Extractor ($\mathcal{E}$): Captures feature maps from a block, analogous to mRNA encoding in biology. This can involve channel concatenation, aggregation, or skip connections.
(2) Transfer ($\mathcal{T}$): Projects extracted features to match the dimension of the receiving block using 1$\times$1 convolutions and (bi)linear resizing. This mimics tRNA in biological systems, aligning representations for compatibility.
(3) Mixture ($\mathcal{M}$): Fuses incoming and native signals, similar to how rRNA coordinates final protein synthesis. This block integrates inherited features while preserving native activations.
\\
Which ETM architecture is optimal clearly varies with the concrete layouts of \parent and \child, but we find that reasonable results can also be obtained with a general purpose architecture.
Given signals $x$ (from the parent) and $x'$ (from the child), the ETM elements proceed as:
    \begin{align*}\text{Extractor: } \mathcal{E}(x) = z \end{align*}
    \begin{align*} \text{Transfer: } \tau = \mathcal{T}(z) \end{align*}
    \begin{align*} \text{Mixture: } \ \phi = \mathcal{M}(x', \tau) \end{align*}
This pipeline operates across each aligned pair $(\parentblock_i, \childblock_i)$, passing $\phi$ forward to the next child layer.


\subsection{Genetic Attention (GA)}
We propose GA as the core mechanism within the Mixture block $\mathcal{M}$, designed to selectively extract complementary information from the parent signal $\tau(x)$ relative to the child’s current representation $x'$. Rather than computing attention as a similarity aggregator, GA acts as a \emph{dissimilarity operator}, highlighting only what the child is missing. In this formulation, we fix the query $Q = x'$ (child activation), and set $K = V = \tau(x)$ (transferred parent representation). Flattening any possibly high dimensions (subscript $_\text{flat}$), we compute:
\begin{equation} 
\alpha = \text{softmax}\left(\frac{Q_{\text{flat}} K_{\text{flat}}^T}{\sqrt{c}}\right), 
\textbf{GA}(\tau, x') = V - \text{reshape}( \alpha V_{\text{flat}} )
\end{equation}
This operation first computes similarity between the child and the parent feature maps. The resulting weighted average of parent features, $\text{reshape}(\alpha V_{\text{flat}})$, represents what the child already perceives. Subtracting it from $V = \tau(x)$ isolates the residual information: the unlearned or underrepresented aspects in the child’s view. The output of the Mixture block is then computed via residual fusion as:
\begin{equation} 
\mathit{ETM}(x, x') = x' + \lambda \cdot \textbf{GA}(\tau(x), x')
\end{equation}
where $\lambda \in [0, 1]$ controls the contribution of the inherited signal. This encourages the child network to internalize only complementary knowledge, while preserving its native representations.

\subsection{Backbone Inheritance}

At each aligned block pair $(\parentblock_i, \childblock_i)$ from Figure \ref{fig:hkt} \circled{a}, we apply the ETM triad to enhance the child representation:
\begin{equation}
z_i^* = \mathit{ETM}(\tilde z_i, z_i)
\end{equation}
This fused output $z_i^*$ is used as the input for subsequent layers in the child network, progressively integrating parent-derived complementary information into the backbone. To enforce learning from the inherited pathway, we supervise the final enhanced output using a mean absolute error loss:
\begin{equation}
\mathscr{L}_1 = \textbf{MAE}(z_n^*, \mathcal{G}_T)
\end{equation}
We make sure the child still learns independently and does not become overly reliant on the parent signal, by simultaneously supervising the native output $z_n$ (see Figure \ref{fig:hkt} \circled{b}) with the task loss:
\begin{equation}
\mathscr{L}_2 = \ell(z_n, \mathcal{G}_T)
\end{equation}
Together, these two losses form a dual supervision strategy: one through inherited, knowledge-enriched features, and the other through self-learned representations. This balances the benefits of guidance and autonomy, supporting both alignment with the parent model and the preservation of architectural efficiency in the child.

\subsection{Head Inheritance}

While the backbone inheritance mechanism enriches early and intermediate representations in the child, the final steps of learning (Figure \ref{fig:hkt} \circled{c}) benefit from leveraging the decision logic encoded in the parent’s output layers. To do this, we reverse the direction of knowledge transfer: the intermediate representations of the child are injected into the parent network using the ETM process.

Formally, at each final stage $i$, we pass $z_{i-1}$ and $\tilde z_{i-1}$ into the ETM function to construct an input to $\parentblock_i$:
\begin{equation}
\tilde z_i = \parentblock_i( \mathit{ETM}(z_{i-1}, \tilde z_{i-1}) )
\end{equation}
This setup forces the child backbone to generate features that are not only optimized for its own head but are also compatible with the parent’s head.
The approach is called head inheritance since the child head is virtually replaced by the parent's head (Figure \ref{fig:hkt} \circled{c}).
This additional training signal promotes latent alignment between the two models. To enforce this alignment, we supervise the parent’s final output using another mean absolute error loss:
\begin{equation}
\mathscr{L}_3 = \textbf{MAE}(\tilde z_n^*, \mathcal{G}_T)
\end{equation}
This loss ensures that the child learns to produce representations that yield accurate predictions when processed by the parent’s decision-making layers, without modifying the parent’s parameters.

\subsection{Combined Loss}
The three pathways described—backbone inheritance, native supervision, and head inheritance—are integrated into a single composite loss:
\begin{equation}
\mathscr{L}_{\text{HKT}} = \alpha_1 \mathscr{L}_1 + \alpha_2 \mathscr{L}_2 + \alpha_3 \mathscr{L}_3
\end{equation}
Here, $\alpha_1, \alpha_2, \alpha_3 \in [0, 1]$, with $\max\{\alpha_1, \alpha_3\} \leq \alpha_2$ and $\sum{\alpha_i}=1$, are manually chosen weighting coefficients to reflect the importance of each learning pathway. This structure enables a flexible, yet disciplined optimization process that balances inherited guidance with child autonomy. The resulting model combines high efficiency with strong task performance, aligned with the design goals of biologically inspired inheritance in deep networks.

\subsection{Training Protocol and Convergence}
During training, only the child model’s parameters $\theta_\mathcal{C}$ are updated; the parent model $\mathcal{P}$ remains frozen and serves as a fixed source of structured knowledge. For each training batch, forward passes are executed through both $\mathcal{P}$ and $\mathcal{C}$ to produce intermediate activations across all aligned blocks. These intermediate signals are then aligned through the transfer block $\mathcal{T}$, integrated via the mixture block $\mathcal{M}$ using the GA mechanism, and propagated within the child network. The resulting outputs are evaluated using the three-stage objective, and gradients are backpropagated solely to update $\theta_\mathcal{C}$ (see Algorithm \ref{alg:3hkt}).

This architecture-agnostic transfer mechanism enables the child network to acquire complementary capabilities from the parent in a modular, interpretable fashion, promoting better generalization under resource constraints. In this work, we study two specific HKT variants: 2-stage HKT (2HKT) and 3-stage HKT (3HKT). 
In this nomenclature, $nHKT$ refers to the process with $n$ functional blocks in the parent and child architectures, that is, the aligned parent-child module pairs used in the transfer process.
Importantly, they do not correspond to the number of loss terms, but rather reflect the structural depth and granularity of knowledge integration across the network hierarchy during training.

\begin{algorithm}[ht!]
\caption{General $n$-stage HKT Training Routine}
\label{alg:3hkt}
\scriptsize
\begin{algorithmic}[1]
    \State \textbf{Input:} $\mathcal{P} = \{p_1, \dots, p_n\}, \ \mathcal{C} = \{c_1, \dots, c_n\}, \ \mathcal{D}$ 
    \State \textbf{Output:} $\theta_{\mathcal{C}}$
    \For{each $(x, \mathcal{G_T}) \in \mathcal{D}$}
        \State $\tilde{z}_1 = p_1(x)$, \ $z_1 = c_1(x)$
        \For{$i = 1 \ \textbf{to} \ n-1$}
            \State $\phi_p^{(i)} = \tilde{z}_i + \lambda \ \textbf{GA}(\tau_p^{(i)}(z_i), \tilde{z}_i)$
            \State $\phi_c^{(i)} = z_i + \lambda \ \textbf{GA}(\tau_c^{(i)}(\tilde{z}_i), z_i)$
            \State $z_{i+1}^* = c_{i+1}(\phi_c^{(i)})$
            \State $\tilde{z}_{i+1}^* = p_{i+1}(\phi_p^{(i)})$
            \State $\tilde{z}_{i+1} = \tilde{z}_{i+1}^*, \ z_{i+1} = z_{i+1}^*$
        \EndFor
        \State $\mathscr{L}_1 = \text{MAE}(z_n^*, \mathcal{G_T})$
        \State $\mathscr{L}_2 = \ell(z_n, \mathcal{G_T})$
        \State $\mathscr{L}_3 = \text{MAE}(\tilde{z}_n^*, \mathcal{G_T})$
        \State $\mathscr{L} = \sum_{j=1}^3 \alpha_j \mathscr{L}_j$
        \State Update $\mathcal{C}$ using $\nabla \mathscr{L}$
    \EndFor
    \State \Return $\theta_{\mathcal{C}}$
\end{algorithmic}
\end{algorithm}
\section{Experiment on Optical Flow Tasks}\label{sec:experiment}
To evaluate the effectiveness of our approach, we started conducting experiments on the optical flow task.
Optical flow estimation involves calculating the dense motion field \(\boldsymbol{F = f(I_a, I_b)}\) between two consecutive images \(\boldsymbol{I_a, I_b \in \mathbb{R}^{H\times W\times 3}}\), where \(\boldsymbol{F \in \mathbb{R}^{H \times W \times 2}}\) represents pixel-wise velocity vectors in a dynamic scene \cite{Fleet2006OpticalFE}. While traditional methods formulated this task as an energy minimization problem under constraints like brightness constancy and motion smoothness \cite{horn1981determining,brox2004high}, deep learning techniques \cite{meister2018unflow,ahmadi2016unsupervised,ranjan2017optical,DBLP:journals/corr/abs-2003-12039} have dramatically advanced performance. However, as demonstrated in Table \ref{tab:performance_general}, these improvements often come at the cost of increased model complexity, making deployment on resource-constrained devices challenging.

We employ a 3HKT framework within a very established modular framework dubbed RAFT \cite{teed2020raft}. The results demonstrate that our method achieves superior performance compared to baseline models and existing knowledge distillation techniques such as DRAFT \cite{10648058}, while also exhibiting robust generalization across multiple benchmark datasets. Additionally, we performed an extensive ablation study using a 2-stage HKT framework. The findings highlight two key insights: (a) increasing the modularization of a model enhances the scope of transferable knowledge, and (b) the proposed methodology delivers consistent and significant improvements in overall performance and effectiveness.
\paragraph{\textbf{HKT for Efficient Optical Flow Estimation:}}
 In our approach, RAFT serves as the parent model, and RAFT-Small (RAFT-S) as the child model. Both networks are modularized into three components to enable staged knowledge transfer: (i) the feature extraction functional block, (ii) the context extraction function block, and (iii) the flow prediction block. The transfer process is guided by the ETM mechanism, which aligns intermediate representations of the parent and child models, facilitating the inheritance of knowledge across modules. The final loss is computed and backpropagated through RAFT-S, refining its parameters with knowledge inherited from RAFT. This method yields 3HKT-RAFT, a compact and efficient model that maintains high accuracy while significantly reducing computational demands, demonstrating the effectiveness of HKT for optical flow tasks.
\paragraph{Experiment Setup: }Experiments were conducted in PyTorch, using a RAFT-inspired setup. Models were trained on datasets FlyingChairs \cite{Dosovitskiy2015FlyingChairs}, FlyingThings3D \cite{Mayer2016FlyingThings3D}, Sintel \cite{10.1007/978-3-642-33783-3_44}, and KITTI-2015 \cite{Menze_2015_CVPR}. Models were pretrained on FlyingChairs and FlyingThings3D, standard benchmarks for optical flow initialization. Generalization performance was evaluated on Sintel and KITTI-2015 training datasets. To enhance robustness and accuracy, fine-tuning was performed on a combined dataset of HD1K \cite{7789500}, Sintel, and KITTI, improving adaptability to diverse motions and environments.
Performance was assessed using End-Point-Error (EPE) and F1-all metrics, following established optical flow benchmarks. EPE was computed on Sintel (clean and final) with validation on a subset of the training split, while F1-all was used for KITTI. Lower values indicate better performance.
\begin{table*}[ht!]
\centering
\scriptsize
\begin{adjustbox}{max width=\textwidth}
\begin{tabular}{llllllllllllllll}
\toprule
\multirow{2}{*}{\textbf{Method}} 
& \multicolumn{4}{c}{C+T} 
& \multicolumn{4}{c}{C+T+S/K} 
& \multicolumn{4}{c}{C+T+S+K+H} 
& KITTI Test & Params & Inf. \\
\cmidrule(lr){2-5} \cmidrule(lr){6-9} \cmidrule(lr){10-13}
\cmidrule(lr){14-14} \cmidrule(lr){15-15} \cmidrule(lr){16-16}
& S-C$\downarrow$& S-F$\downarrow$ & K-E$\downarrow$ & K-All$\downarrow$ 
& S-C$\downarrow$ & S-F$\downarrow$ & K-E$\downarrow$ & K-All$\downarrow$ 
& S-C$\downarrow$ & S-F$\downarrow$ & K-E$\downarrow$ & K-All$\downarrow$ 
& F1-All$\downarrow$ & (M)$\downarrow$ & (ms)$\downarrow$ \\
\midrule

PWC-Net~\cite{sun2018pwc} & 2.55 & 3.93 & 10.35 & 30.7 & - & - & - & - & - & - & - & - & 9.60 & 9.4 & 30 \\
LiteFlowNet/2~\cite{hui2018liteflownet,hui2020lightweight} & 2.48 & 4.04 & 10.39 & 28.5 & - & - & - & - & (1.30) & (1.62) & (1.47) & (4.8) & 7.74 & 4.8 & 139 \\
FlowID~\cite{xu2021high} & 1.98 & 3.27 & 6.69 & 22.95 & - & - & - & - & - & - & - & - & 6.27 & 5.7 & 332 \\
GMFlow~\cite{DBLP:journals/corr/abs-2111-13680} & 1.08 & 2.48 & 7.77 & 23.4 & - & - & (2.85) & (10.77) & - & - & - & - & 9.32 & \underline{4.7} & 156 \\
FlowFormer~\cite{huang2022flowformer} & \textbf{0.64} & \textbf{1.5} & \textbf{4.09} & \textbf{14.72} & - & - & - & - & - & - & - & - & \textbf{4.68} & 18.2 & 970 \\
GMFlowNet & 1.14 & 2.71 & 4.24 & 15.4 & - & - & - & - & 1.39 & 2.65 & 4.79 & - & - & 9.3 & - \\
VCN~\cite{yang2019volumetric} & 1.30 & 2.59 & 4.60 & 15.9 & (1.66) & (2.24) & (\textbf{1.16}) & (\textbf{4.10}) & - & - & - & - & 6.30 & 6.2 & \textbf{26} \\
\midrule
RAFT (\(\mathcal{P}\))~\cite{DBLP:journals/corr/abs-2003-12039} & 1.43 & 2.71 & 5.04 & 17.47 & (\textbf{0.77}) & (\textbf{1.22}) & (2.45) & (7.90) & (\textbf{0.64}) & (\textbf{1.22}) & (\textbf{0.63}) & (\textbf{1.5}) & \underline{5.10} & 5.3 & 393 \\
RAFT (\(\mathcal{C}\)) Baseline & 2.21 & 3.35 & 7.51 & 26.9 & (1.48) & (2.04) & (7.66) & (21.75) & (1.48) & (2.14) & (2.43) & (10.47) & 10.77 & \textbf{1.0} & \underline{50} \\
DRAFT~\cite{10648058} & 1.99 & 3.17 & 6.94 & 23.92 & (1.32) & (1.89) & (5.04) & (17.14) & (1.94) & (2.23) & (9.45) & (10.35) & 7.53 & \textbf{1.0} & \underline{50} \\
\midrule
2HKT-RAFT (Ours) & 1.91 & \underline{3.03} & 7.37 & 26.11 & (1.18) & (1.82) & (\underline{4.58}) & (\underline{16.39}) & (1.20) & (1.79) & (2.09) & (8.65) & - & \textbf{1.0} & \underline{50} \\
\textbf{3HKT-RAFT (Ours)} & \underline{1.90} & 3.08 & \underline{6.45} & \underline{24.73} & (\underline{1.12}) & (\underline{1.62}) & (4.94) & (16.49) & (\underline{1.13}) & (\underline{1.63}) & (\underline{1.96}) & (\underline{8.03}) & - & \textbf{1.0} & \underline{50} \\
\bottomrule
\end{tabular}
\end{adjustbox}
\vspace{-0.6\baselineskip}
\caption{3HKT-RAFT improves accuracy, compression, and inference speed over baselines and many SOTA approaches. Bold = best, underline = second-best. On training sets: S-C, S-F, K-E, K-All = Sintel clean, Sintel Final, Kitti F1-EPE, Kitti F1-All. We carry out our experiment on TITAN RTX TU102.}
\label{tab:performance_general}
\end{table*}

\subsection{Results}
\textbf{Accuracy and Generalization ---} Table \ref{tab:performance_general} summarizes the performance of 2-stage and 3-stage HKT across optical flow benchmarks after training on FlyingChairs and FlyingThings3D (C+T) and with Sintel and KITTI data augmentation (C+T+S/K). When trained on C+T, 3HKT consistently achieves lower EPE compared to 2HKT, particularly on challenging datasets such as Sintel (clean:$1.9$) and KITTI (F1-all: $24.73$). This demonstrates the superior ability of 3HKT to estimate fine-grained motion and generalize in diverse scenarios. Due to limited computational resources, training on C+T+S/K was conducted for only half the number of epochs used for 2HKT. Even with reduced training, 3HKT demonstrates competitive results, achieving significantly better accuracy than RAFT-S, though 2HKT performed marginally better in this scenario.

\paragraph{Efficiency: Model Compression and Inference Speed---}In addition to its accuracy, 3HKT delivers substantial improvements in efficiency. By leveraging the modular ETM triad and enhanced GA, 3HKT reduces the parameter count by \textbf{80\%} compared to the parent model while achieving a \textbf{6×} inference speedup. This efficiency, combined with its competitive performance, makes 3HKT highly suitable for deployment in resource-constrained environments, such as edge devices and real-time systems.
\paragraph{Scalability to Multi-Stages ---}The performance trends in Table \ref{tab:performance_general} highlight the scalability potential of 3HKT for multi-stage frameworks. The addition of the third stage improves task-specific knowledge integration without introducing significant computational overhead. This supports the hypothesis that extending HKT to more stages can further enhance adaptability for complex modular architectures and diverse AI applications. These findings demonstrate that 3HKT not only outperforms RAFT-S and offers competitive performance relative to 2HKT but also establishes a robust foundation for scalable multi-stage inheritance in broader AI contexts, paving the way for modular and efficient knowledge transfer frameworks.
\paragraph{Qualitative Experiment Results ---}Figure \ref{fig:figure3} illustrates the qualitative advantages of 2-stage and 3-stage HKT over RAFT-S and DRAFT (a KD approach to RAFT compression) on synthetic (Sintel) and real-world (KITTI) datasets. Both 2HKT and 3HKT effectively capture finer motion details and delineate object boundaries, particularly in regions with complex textures, occlusions, or dynamic motion patterns. The additional inheritance stage in 3HKT improves the fusion of knowledge between the models of parents and children, enabling precise modeling of intricate motion while leveraging the GA mechanism to dynamically prioritize the relevant features of the task. However, 3HKT's multi-stage design introduces trade-offs, such as sensitivity to noise in low-texture regions, where feature extraction may falter. These results underscore the robustness of 3HKT for challenging motion estimation tasks, while also highlighting the balance required between refinement and efficiency.
\begin{figure}[ht!]
    \centering
    \includegraphics[width=\linewidth]{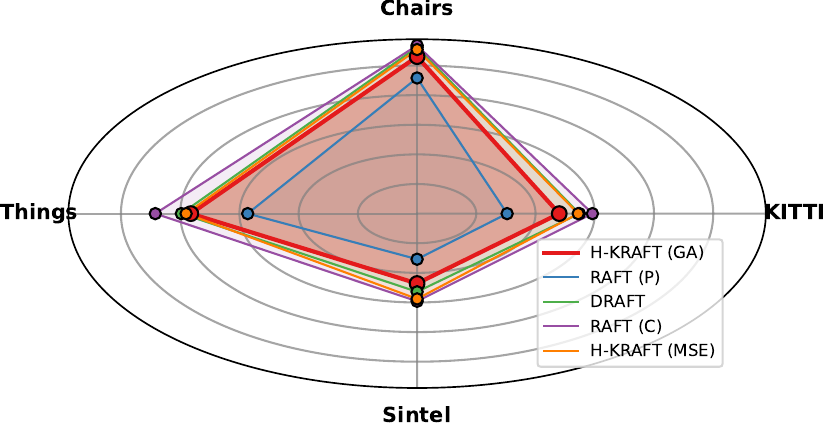}
    \vspace{-1.5\baselineskip}
    \caption{\textbf{Generalization Performance}: H-KRAFT (3HKT-RAFT with $\circ$=\textbf{GA} or  \textbf{MSE}) achieves superior generalization with the best performance ($\mu$) across datasets compared baseline models.}
    \label{fig:eval}
\end{figure}
\paragraph{Ablation Studies: }Table \ref{fig:eval} quantifies the contributions of the third stage, GA, and structural units within 3HKT, using the aggregate metric $\mu = \text{AVG}(\text{metrics})$ across benchmarks. Here, values of $\mu$ closer to the center indicate better performance. The third stage boosts task-specific transfer, improving $\mu$ by $11\%$, reducing Sintel (clean) EPE by $14\%$, and raising KITTI F1-all by $14\%$. Removing GA lowers $\mu$ by $8\%$ and increases Sintel (clean) EPE by $11\%$. Structural ablations confirm extractor and transfer units enable precise alignment, while the mixture unit is crucial—its removal raises EPE by $31\%$ (computed as here\footnote{$\circ = \text{MSE}$, with $\text{MSE}(\tau^{(i)}, z_i) = \frac{1}{n}\sum_{j=1}^{n} (\tau^{(i)}_j - z_{i,j})^2$}). The inheritance of the backbone and head further supports the performance, confirming the reliability and scalability of 3HKT-RAFT.
\section{Experiment on Image Classification Tasks}
To evaluate the generalizability of the HKT framework beyond optical flow, we extend our investigation to image classification using the CIFAR-10 dataset \cite{krizhevsky2014cifar}. We adopt ResNet \cite{He_2016_CVPR} architectures, known for their modular design and strong performance, making them ideal candidates for our heredity-based transfer paradigm. Specifically, we select ResNet-110 as the parent model $\mathcal{P}$ and ResNet-20 as the child model $\mathcal{C}$. 
\begin{table}[ht!]
    \centering
    \setlength{\tabcolsep}{2pt} 
    \renewcommand{\arraystretch}{1} 
    \begin{tabular}{lcccc}
        \toprule
        \textbf{Model} & \textbf{Prec. (\%)}$\uparrow$ & \textbf{Error (\%)}$\downarrow$ & \textbf{Params (M)}$\downarrow$ & \textbf{Inf. (ms)}$\downarrow$ \\
        \midrule
        $\mathcal{P}$ & \textbf{93.6}   & \textbf{6.43} & 1.7 & $\sim$0.30 \\
        $\mathcal{C}$ & 91.25           & 8.75          & \textbf{0.27} & \textbf{$\sim$0.10} \\
        \textbf{Ours} & \underline{92.4} & \underline{7.6} & \textbf{0.27} & \textbf{$\sim$0.10} \\
        \bottomrule
    \end{tabular}
    \vspace{-0.6\baselineskip}
    \caption{Evaluation of ResNet-20 ($\mathcal{C}$), ResNet-110 ($\mathcal{P}$) and HKT-ResNet-20 (Ours) on CIFAR-10. HKT improves the child’s performance while maintaining compactness. Experiments on TITAN RTX TU102.}
    \label{tab:resnet_evaluation}
\end{table}
\paragraph{Experiment Setup ---}Following the procedure outlined in Algorithm \ref{alg:3hkt}, we modularize both ResNet-110 and ResNet-20 into two functional stages for inheritance. These stages correspond to the output of the average pooling layer (feature abstraction) and the final fully connected (FC) layer (prediction). These layers serve as inheritance points in our 2-stage HKT setup. Given the architectural consistency across ResNet variants, no explicit transformation module ($\mathcal{T}$) is required, and extraction ($\mathcal{E}$) is directly realized through the forward pass. Thus, inheritance is governed mainly by the Mixture module ($\mathcal{M}$) from the ETM triad. We follow the standard 2HKT training procedure, using CIFAR-10 as the benchmark dataset. The child model is optimized using an aggregated loss combining core supervision with backbone and head inheritance components. 

\paragraph{Results ---}As shown in Table \ref{tab:resnet_evaluation}, HKT-ResNet-20 achieves a top-1 precision of \underline{92.4\%}, improving upon the baseline ResNet-20 (91.25\%) while preserving its compactness (0.27M parameters). This corresponds to an 8.8\% relative error reduction, narrowing the gap with the deeper ResNet-110 parent (93.6\%). Inference remains efficient: while ResNet-110 requires $\sim$0.30 ms per CIFAR-10 image on a TITAN RTX TU102, both the baseline ResNet-20 and HKT-ResNet-20 maintain $\sim$0.10 ms per image. These results demonstrate that HKT enables small classification models to inherit semantic precision from larger networks, achieving a superior balance between accuracy and efficiency.  
               
\section{Experiment on Image Segmentation Tasks}
To further explore the generalizability of HKT, we extend our evaluation to semantic segmentation, applying the framework to U-Net architectures—a well-established family of models in this domain. Specifically, we select PocketNet (Mini U-Net) as the child model $\mathcal{C}$ \cite{celaya2022pocketnet} and H-DenseUNet \cite{li2018hdenseunet} as the parent model $\mathcal{P}$, both evaluated on the LiTS dataset \cite{bilic2023liver} for liver segmentation. 

\paragraph{Experiment Setup ---}In accordance with Algorithm \ref{alg:3hkt}, we modularize both networks into aligned stages and apply a 3-stage HKT procedure. The functional blocks used for inheritance correspond to key encoding and decoding stages in the U-Net pipeline, including intermediate feature abstraction and the final segmentation output. Given the architectural consistency between the two networks, no additional transformation ($\mathcal{T}$) is required, and feature extraction ($\mathcal{E}$) is directly realized through existing encoder outputs. Thus, the inheritance is governed mainly by the Mixture $\mathcal{M}$ module, implemented via Genetic Attention.
\begin{table}[ht!]
    \centering
    \begin{tabularx}{\linewidth}{Xccc}
        \toprule
        \textbf{Model} & \textbf{D-Mean}$\uparrow$ & \textbf{Inf. (s)}$\downarrow$& \textbf{Params (M)}$\downarrow$ \\
        \midrule
        $\mathcal{P}$    & \textbf{96.5$\%$}   & 30--200   & 80 \\
        $\mathcal{C}$    & 93.0$\%$            & 88--99    & \textbf{0.8} \\
        \textbf{Ours}    & \underline{94.1$\%$}& \underline{88--99}    & \textbf{0.8} \\
        \bottomrule
    \end{tabularx}
    \vspace{-0.6\baselineskip}
    \caption{Evaluation of PocketNet / Mini U-Net ($\mathcal{C}$), H-DenseUNet ($\mathcal{P}$), and HKT-U-Net (Ours) on the LiTS dataset. HKT improves the child’s performance while maintaining compactness and efficiency. Experiments on TITAN RTX TU102.}
    \label{tab:unet_evaluation}
\end{table}

\begin{figure*}
    \centering
    \begin{subfigure}[b]{\linewidth}
        \includegraphics[width=\linewidth]{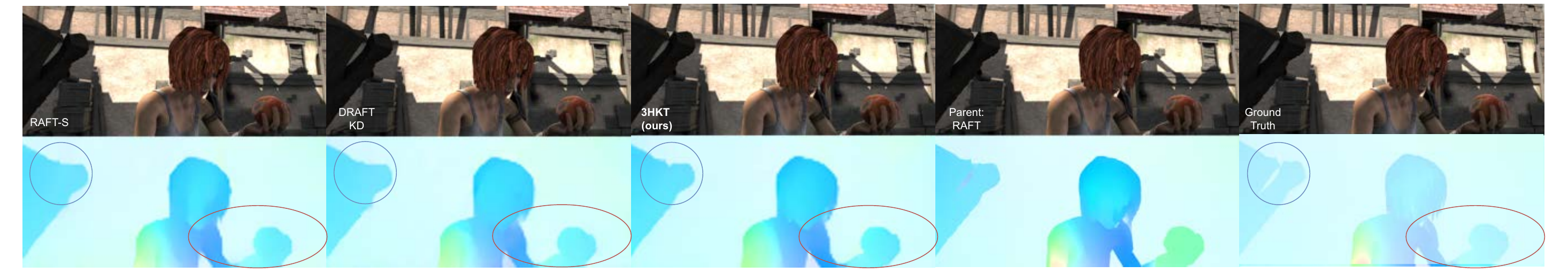}
    \end{subfigure}
    \begin{subfigure}[b]{\linewidth}
        \includegraphics[width=\linewidth]{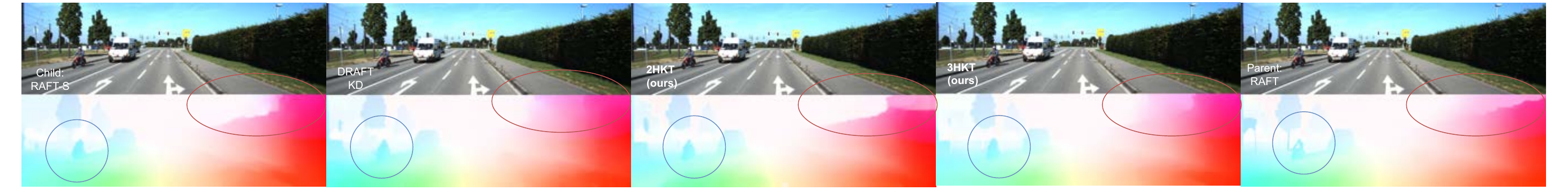}
    \end{subfigure}  
    \begin{subfigure}[b]{\linewidth}
        \includegraphics[width=\linewidth]{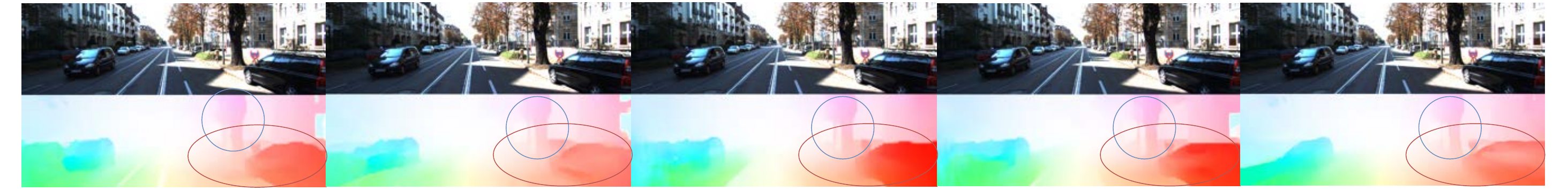}
    \end{subfigure}
    \vspace{-1.5\baselineskip}
    \caption{\textbf{Qualitative Comparison:} Highlighted regions demonstrate that 2-stage and 3-stage HKT effectively capture fine-grained anatomical details and lesion boundaries that baseline models fail to reproduce.}
    \label{fig:figure3}
\end{figure*}
\paragraph{Results ---}The evaluation, as reported in Table \ref{tab:unet_evaluation}, shows that the HKT-enhanced PocketNet achieves a global mean Dice (D-Mean) of $94.1\%$, approaching the H-DenseUNet’s performance ($96.5\%$) while retaining the child’s efficiency. Compared to the baseline PocketNet (D-Mean $93.0\%$), HKT delivers a consistent improvement with negligible inference cost increase. Importantly, PocketNet achieves these results with only $0.8$M parameters—over $100\times$ smaller than the $80$M parameter H-DenseUNet—while reducing inference time from $30$–$200$ seconds per CT scan (H-DenseUNet) to $88$–$99$ seconds. These results confirm that HKT enables compact segmentation models to inherit semantic precision from larger architectures, providing performance close to state-of-the-art full-capacity models without sacrificing deployability.

\section{Limitations and Future Work}\label{sec:lim&future}
While HKT demonstrates consistent improvements in accuracy, generalization, and model compression, its design introduces practical trade-offs that must be considered when applying it to different domains and architectures.
\\
\textbf{Training Overhead:} One of the primary limitations of HKT is the increase in training time. While inference remains efficient—thanks to the compact, standalone child architecture—training requires both the parent and child to remain active, with forward passes through aligned blocks and additional computation for the ETM triad and Genetic Attention. Specifically, we observe that 3HKT requires approximately $1.74\times$ more training iterations than 2HKT, and around $3\times$ more than baseline knowledge distillation methods. Although GA is lightweight and operates on intermediate features, the added third inheritance stage and the fusion overhead contribute to a measurable increase in training duration. This does not imply faster convergence, as the child model must now optimize both native and complementary inherited features, which adds complexity to the learning dynamics.
\\
\textbf{Modular Scalability:} Another challenge lies in the assumption of architectural modularity. The multi-stage inheritance process relies on functional alignment between blocks of the parent and child networks. This alignment is natural in architectures like U-Net, RAFT and ResNet, but becomes difficult when compressing heterogeneous or fused models. For instance, transferring from a transformer-based model to a convolutional architecture introduces ambiguity in mapping semantic functionality across stages. This limits HKT's out-of-the-box applicability in settings with weak structural correspondence.
\\
\textbf{Future Directions:} To overcome these challenges, future work will explore dynamic block matching techniques using (dis)similarity-aware alignment or graph-based analysis of activation patterns. Further improvements to the ETM and GA components, such as selective gating, adaptive stage weighting, or sparsification, may reduce overhead and enhance transfer precision. Extending HKT to continual learning, unsupervised settings, or cross-domain tasks (e.g., segmentation, depth, or language models) represents another promising direction toward broader generalization and real-world impact.

\section{Conclusion}\label{sec:conclu}
This paper introduced HKT, a biologically inspired framework for modular, stage-wise neural inheritance. Built on the ETM triad and Genetic Attention mechanism, HKT enables selective, semantically aligned knowledge transfer from a larger parent network to a compact child model. Evaluations on optical flow benchmarks (Sintel, KITTI), image classification (CIFAR-10) and semantic segmentation (LiTS) confirm that HKT significantly outperforms state-of-the-art distillation approaches in both accuracy and compression, with the 3-stage variant yielding the best overall alignment and predictive robustness. HKT offers a scalable, task-agnostic strategy for deep model compression, bridging biological principles with modern neural architecture design.

\bibliography{aaai2026}

\end{document}